\documentclass[12pt]{amsart}
\usepackage{times}
\usepackage{amssymb, amsmath}
\usepackage{amsthm}
\usepackage{tikz}
\usepackage{bm}
\usetikzlibrary{matrix,arrows}
\usepackage{enumitem}
\usepackage{booktabs, float}
\usepackage[margin=2.5cm]{geometry}

\theoremstyle{plain}

\theoremstyle{remark}

\begin{document}

\title[Partial Resampling of Imbalanced Data] {Partial Resampling of Imbalanced Data}
\author[F. Kamalov, A. Atiya, D. Elreedy]{Firuz Kamalov$^1$$^{\boldsymbol{*}}$, Amir F. Atiya$^2$, Dina Elreedy$^3$}

\address{$^{1}$ Canadian University Dubai, Dubai, UAE.}
\email{\textcolor[rgb]{0.00,0.00,0.84}{firuz@cud.ac.ae}}

\address{$^{2}$ Cairo University, Cairo, Egypt}
\email{\textcolor[rgb]{0.00,0.00,0.84}{amir@alumni.caltech.edu}}

\address{$^{3}$ Cairo University, Cairo, Egypt.}
\email{\textcolor[rgb]{0.00,0.00,0.84}{dinaelreedy@email.wustl.edu}}

%\author{Firuz Kamalov}
%\subjclass[2000]{6207}
%^\address{Mathematics Department, Canadian University of Dubai, Dubai, UAE}
%\email{firuz.kamalov@huskers.unl.edu}
%\date{\today}
\date{\today
\newline \indent $^{\boldsymbol{*}}$ Corresponding author
%\newline \indent Preprint of an article submitted for consideration in Annals of Data Science
}

\begin{abstract}

Imbalanced data is a frequently encountered problem in machine learning. Despite a vast amount of literature on sampling techniques for imbalanced data, there is a limited number of studies that address the issue of the optimal sampling ratio. In this paper, we attempt to fill the gap in the literature by conducting a large scale study of the effects of sampling ratio on classification accuracy. We consider 10 popular sampling methods and evaluate their performance over a range of ratios based on 20 datasets. The results of the numerical experiments suggest that the optimal sampling ratio is between 0.7 and 0.8 albeit the exact ratio varies depending on the dataset. 
Furthermore, we find that while factors such the original imbalance ratio or the number of features do not play a discernible role in determining the optimal ratio, the number of samples in the dataset may have a tangible effect. 

\end{abstract}

\maketitle

\section{Introduction}
Imbalanced data refers to the skewed distribution of class labels in data. It occurs in a wide range of fields including medicine \cite{Yildirim}, cybersecurity \cite{Sun}, fraud detection \cite{Hassan}, and others.
One of the common approaches to dealing with imbalanced data is through sampling. In most cases, sampling is applied until a fully balanced dataset - where each class has an equal number of samples - is achieved. The balanced dataset is then passed on to the training phase \cite{Buda}. However, it is not necessarily true that the training set must be fully balanced to achieve the optimal results.

The traditional argument for balancing the data is that classifiers tend to focus on the majority class samples at the expense of the minority class. For instance, given a training set with a 1/100 class imbalance ratio, most classifiers would end up classifying all the samples as negative (majority). Therefore, it is argued that class balancing is required to force classifiers to learn the minority samples. On the other hand, by artificially balancing the data we are distorting the reality. Ideally, the class ratios in the training and testing set should be the same to maintain the fidelity of the process. Sampling the training set while keeping the test set at the original class ratio goes against this philosophy of congruent train and test sets. 

Excessive sampling - where the sampled class ratio is far greater than the original class ratio - may  lead to artificial bias towards the minority class. In most cases, the increased accuracy on the minority class is accompanied with a drastic decrease in accuracy on the majority class. While in some instances the accuracy on minority class is more important than the majority class - as in the case of fraud detection or medical diagnostics - it is worthwhile to take into account the accuracy on the majority class. Therefore, the optimal sampling ratio of the minority to majority samples must be chosen with great care.

Increasing the number of minority points through ovesampling provides the classifier with more data to learn the representation of the minority class. On the other hand, since the new minority points are not generated from the true distribution, the increase in the number of sampled points may lead to model misspecification \cite{Elreedy2019}. At some stage, the issues related to model misspecification outweight the benefits of learning the class representations. The optimal sampling ratio occurs at the stage when the model accuracy begins to deteriorate.

A compromise between full sampling and the original class ratio is partial sampling. Partial sampling implies sampling the data up to a specified class ratio (minority/majority) that is between the original class ratio and the 50/50 ratio. To obtain the optimal partial sampling ratio a grid search procedure can be utilized. Concretely, we can measure the performance of a classifier on the training set for different partial sampling ratios using cross-validation. After identifying and training with the optimal partial sampling ratio, the classifier is tested on the holdout set.

In this paper, we conduct a large-scale, systematic study of the effects of partial sampling of imbalanced data on the performance of classifiers. In particular, we evaluate the accuracy of classifiers that are trained on datasets sampled over a range of class ratios.
Our study is distinguished from other similar studies by its breadth. While other studies employ only a few datasets and sampling methods \cite{Buda, Seo, Thabtah}, we consider 20 imbalanced datasets (Table \ref{datasets}) and 10 sampling methods (Table \ref{sampling}). For each dataset, we sample the original data with different ratios and evaluate the performance of the trained classifier. We use random forest (RF) and support vector machines (SVM) as the base classifiers in our numerical experiments. Given the scale of the present study, its results provide valuable insights into the effects of sampling ratio.

The results of the numerical experiments reveal two key insights: i) the  optimal sampling ratio of the minority to majority samples is almost always less than 1 and often in the range of 0.7-0.8, and ii) the optimal ratio depends on the particular dataset. The results hold across an array of popular sampling techniques. Based on the outcomes of our empirical study, we recommend selecting the optimal sampling ratio using a grid search procedure. Furthermore, the default sampling ratio is recommended to be 0.75. We hope that our study helps researchers to better understand the effects of different sampling ratios and expedite the selection of the optimal ratio.

The paper is structured as follows. Section 2 discuses the current literature regarding imbalanced data and sampling methods. Section 3 presents the methodology employed in the study including the experimental setup, datasets, sampling methods, classifiers, and metrics. Section 4 contains the results and analysis of the numerical experiments.
Section 5 concludes the paper with summary remarks and discussion of future research.
%-----------------------------------------------------------------------------------------------------------------------------------------------------
%-----------------------------------------------------------------------------------------------------------------------------------------------------
\section{Literature}

Data sampling is a widely used pre-processing step in many machine learning pipelines. Despite some skepticism about the usage of sampling \cite{Moniz} most researchers agree about its efficacy. There exists a variety of sampling techniques in the literature. A number of sampling techniques operate by generating new minority points between the existing
neighboring points. For instance, in \cite{Chawla} the new points are generated using a uniform distribution, while in \cite{Kamalov1} the new points are generated via a gamma distribution.  In a localized approach, more points are generated in the regions closer to the majority class  points \cite{Chen, Zhu}. New samples can also be generated by first estimating the underlying distribution of the minority points. Given a learned distribution, the new points are then obtained from the density distribution. To learn the distribution of the points both statistical  and deep learning models are used . In the statistical approach kernel density estimation is applied \cite{Kamalov2}, while in the deep learning approach generative adversarial networks are often employed \cite{Shamsolmoali, Zhang}.

Partial sampling and its effects on classification accuracy have been studied by several authors. The authors in \cite{Weiss} found that the optimal sampling ratio depends on the classifier performance metric. While the full sampling ratio is preferred under the AUC, the original sampling ratio is preferred based on the classification accuracy. In \cite{Albisua}, the authors concluded that the optimal class distribution is not necessarily achieved at the fully balanced distribution.
Similarly, in \cite{Buda} the authors studied the effects of different imbalance ratios in the context of image classification and  convolutional neural networks. Their experiments showed that class imbalance is generally detrimental to classifier performance. They also found oversampling to be the most effective method for combating class imbalance. 
The effects of class imbalance were also studied in \cite{Thabtah} who used undersampling to balance the data. The authors found that classification precision and recall are the lowest at class ratio 0.5. 

The authors of \cite{Garcia} considered two factors affecting the performance of sampling methods: the employed classifier and the degree of imbalance. Their results indicate that the best sampling method depends of the class imbalance ratio. They concluded that for datasets having low or moderate class imbalance ratio, oversampling outperforms undersampling using local classifiers such as kNN. However, some undersampling methods outperform over-sampling when using global learning classifiers such as neural networks.
The authors in \cite{Bonas} used random oversampling and undersampling to evaluate the efficacy of different sampling ratios. The results show no significant difference in classification accuracy for different sampling ratios with only a small decrease around $r=1$.
In \cite{Seo}, the authors seek to find the optimal sampling ratio for intrusion detection data (KDD99). The authors use SMOTE to test the performance of various sampling ratios showing that the class ratio of $r=10$ provides the optimal results.
A large scale study of 85 different oversampling methods was done in \cite{Kovacs2}. Although the primary goal of the study was to identify the best sampling method, the authors used a range of sampling ratios to train the classifier. Unfortunately, the study did not indicate the performance of the sampling methods at different ratios. 
A broader comparison of data-driven, algorithmic, and hybrid approaches was conducted in \cite{Fathy}.

As discussed above, there are a number of studies that explore the issue of the optimal sampling ratio. However, the majority of the studies are either outdated or limited in scope. Our study provides an up-to-date evaluation of the popular sampling algorithms based on a large scale experimental database.

%-----------------------------------------------------------------------------------------------------------------------------------------------------
%-----------------------------------------------------------------------------------------------------------------------------------------------------
\section{Methodology}
In this section, we discuss the experimental setup, datasets, sampling algorithms, and classifiers used in our study.
\subsection{Experimental setup}\label{sampling}
The objective of this study is to evaluate the efficacy of different sampling ratios for imbalanced data. To this end, we consider a number of different imbalanced datasets to which we apply various sampling techniques to achieve a range of class ratios. In particular, we use cross-validation to split the datasets into train and test sets. The train set is sampled to achieve a given class ratio. Then a classifier is trained on the partially balanced data and tested on the holdout set. The classifier hyperparameters are tuned using cross-validation. The performance of the classifier on the holdout set is measured using balanced accuracy and F1-macro.
A detailed description of the experiment is given below. 
\\
\\
\noindent\textbf{The experimental procedure}
\\
{\small
\text{For each dataset:}
\begin{enumerate}[itemsep=2pt, parsep=2pt]
\item {Split the dataset using 4-fold cross-validation.}
\item {For each fold of cross-validation: }
\begin{enumerate}[label=\roman*.]
\item Resample the train set (the remaining three folds) according to a specified ratio (m/M) in the range 0.2 to 1.
\item Tune the classifier on the resampled train set using a separate 4-fold cross-validation. Use balanced accuracy to select the best model hyperparameters.
\item {Run the trained classifier on the test set.}
\item \text{Record classification (balanced) accuracy and F1-macro on the test subset.}
\end{enumerate}
\item \text{Calculate the average (balanced) accuracy and F1-macro over the 4 validation folds.}
\item \text{As a reference, train and test the classifier using the original data without resampling.}
\end{enumerate}
}

All the numerical experiments were carried out in Python using scikit \cite{Pedregosa}, imblearn \cite{imblearn}, and smote-variants \cite{Kovacs1} libraries.

%-----------------------------------------------------------------------------------------------------------------------------------------------------
%-----------------------------------------------------------------------------------------------------------------------------------------------------
\subsection{Datasets}\label{data}
The main drawback of the existing literature on class imbalance ratios is the limited amount of data employed in the studies. In most cases, the studies employ only a few datasets. To fill this gap in the literature, our study uses 20 datasets.
The datasets used in the study are presented in Table \ref{datasets}. The datasets are selected from a wide range of applications including medicine, image recognition, engineering, and others. The class ratios of the datasets range from 8.6:1 to 26:1. Similarly, the number of features and sample size vary considerably providing a broad spectrum of data for our  analysis. All the data used in the study is publicly available though the UCI Machine Learning Repository \cite{Dua}.

\begin{table}[htb]
\centering
\caption{Datasets used in the study.}
\label{datasets}
\begin{tabular}{lllllrr}
\toprule
{} &            Name &          Repository \& Target &  Ratio &      Samples &   Features &      Size \\
ID &                 &                              &        &         &      &           \\
\midrule
1  &           ecoli &             UCI, target: imU &  8.6:1 &     336 &    7 &     20227 \\
2  &  optical\_digits &               UCI, target: 8 &  9.1:1 &   5,620 &   64 &   3273088 \\
3  &        satimage &               UCI, target: 4 &  9.3:1 &   6,435 &   36 &   2154438 \\
4  &      pen\_digits &               UCI, target: 5 &  9.4:1 &  10,992 &   16 &   1653197 \\
5  &         abalone &               UCI, target: 7 &  9.7:1 &   4,177 &   10 &    405169 \\
6  &  sick\_euthyroid &  UCI, target: sick euthyroid &  9.8:1 &   3,163 &   42 &   1301891 \\
7  &    spectrometer &            UCI, target: $\geq$44 &   11:1 &     531 &   93 &    543213 \\
8  &     car\_eval\_34 &    UCI, target: good, v good &   12:1 &   1,728 &   21 &    435456 \\
9  &          isolet &            UCI, target: A, B &   12:1 &   7,797 &  617 &  57728988 \\
10 &        us\_crime &           UCI, target: $\geq$0.65 &   12:1 &   1,994 &  100 &   2392800 \\
11 &       yeast\_ml8 &            LIBSVM, target: 8 &   13:1 &   2,417 &  103 &   3236363 \\
12 &           scene &   LIBSVM, target: $>$one label &   13:1 &   2,407 &  294 &   9199554 \\
13 &     libras\_move &               UCI, target: 1 &   14:1 &     360 &   90 &    453600 \\
14 &    thyroid\_sick &            UCI, target: sick &   15:1 &   3,772 &   52 &   2942160 \\
15 &       coil\_2000 &  KDD, CoIL, target: minority &   16:1 &   9,822 &   85 &  13357920 \\
16 &      arrhythmia &              UCI, target: 06 &   17:1 &     452 &  278 &   2136152 \\
17 &  solar\_flare\_m0 &            UCI, target: M-$>$0 &   19:1 &   1,389 &   32 &    844512 \\
18 &             oil &        UCI, target: minority &   22:1 &     937 &   49 &   1010086 \\
19 &      car\_eval\_4 &           UCI, target: vgood &   26:1 &   1,728 &   21 &    943488 \\
20 &    wine\_quality &       UCI, wine, target: $\leq$4 &   26:1 &   4,898 &   11 &   1400828 \\
\bottomrule
\end{tabular}
\end{table}
%-----------------------------------------------------------------------------------------------------------------------------------------------------
%-----------------------------------------------------------------------------------------------------------------------------------------------------
\subsection{Sampling methods}\label{sampling_tech}
We consider ten different sampling techniques in our study: eight oversampling and two undersampling algorithms. The list of the sampling methods is provided in Table \ref{sampling}. The list includes classical as well as the state-of-the-art algorithms.

\begin{table}[htb]
\centering
\caption{Sampling methods used in the study.}
\label{sampling}
\begin{tabular}{llll}
\toprule
{} &            Name &          Source&  Inclusion criterion  \\
ID &                 &                              &          \\
\midrule
1  &           SMOTE  &             Chawla et al. (2002) \cite{Chawla} &  Popularity  \\
2  &  ADASYN & He et al. (2008) \cite{He} &  Popularity \\
3  &  Borderline SMOTE  & Han et al. (2005) \cite{Han} &  Popularity \\
4  &  SVM SMOTE & Nguyen et al. (2009) \cite{Nguyen} &  Popularity \\
5  &  NearMiss  & Mani et al. (2003) \cite{Mani} &  Popularity \\
6  &  Random oversampling  &  public &  Popularity \\
7  &  Random undersampling  &  public &  Popularity \\
8  &  ProWSyn  & Barua et al. (2013) \cite{Barua} &  Performance \\
9  &  Polynomial SMOTE  & Gazzah et al. (2008) \cite{Gazzah} &  Performance \\
10  &  Lee  & Lee et al. (2015) \cite{Lee} &  Performance \\

\bottomrule
\end{tabular}
\end{table}

Following the lead of the recent empirical study \cite{Kovacs2}, we consider the top-ranked oversampling methods: Polynomial SMOTE \cite{Gazzah}, ProWSyn \cite{Barua}, and Lee \cite{Lee} in our experimental study.
In addition, we include popular oversampling methods: SMOTE \cite{Chawla}, ADASYN \cite{He}, Borderline SMOTE \cite{Han}, SVM SMOTE \cite{Nguyen}, NearMiss \cite{Mani}, and the basic baselines of random oversampling and undersampling. 
Many of these sampling methods are based on the SMOTE framework, where a new minority sample is randomly generated along the straight light connecting a pair of existing minority points.

\subsection{Classifiers and metrics}\label{classifiers}
We use the performance of the classifier trained on resampled data as a proxy for the efficacy of the sampling strategy. In particular, given an imbalanced dataset, we split it into train and test subsets, and resample the train set up to the given class ratio. Then the classifier is trained on the resampled train set and evaluated on the test set. The performance of the classifier is evaluated using balanced accuracy and F1-macro. Balanced accuracy is defined as the unweighted mean accuracy on the majority and minority subsets. Similarly, F1-macro is defined as the unweighted mean F1-score on the majority and minority subsets. Since the goal of resampling is to improve the classification accuracy on the minority samples, the use of balanced accuracy and F1-macro is recommended. Through the remainder of the paper we will refer to balanced accuracy as simply accuracy.

We employ two standard classifiers in our experiments: random forest (RF) and support vector machines (SVM). The two classifiers have also been used in previous studies \cite{Bonas, Seo}. The RF algorithm is a widely used ensemble classifier that is based on aggregating several individual decision tree classifiers into a single learner. The SVM algorithm is another popular classifier that uses the kernel trick to learn a nonlinear decision boundary. We also considered using a deep neural network as the third base classifier but it is computationally infeasible given the number of experiments conducted in our study.

%-----------------------------------------------------------------------------------------------------------------------------------------------------
%-----------------------------------------------------------------------------------------------------------------------------------------------------
\section{Results and analysis}
In this section, we present and discuss the results of our numerical experiments aimed at understanding the effects of different sampling ratios. The results are based on evaluation of 20 imbalanced datasets, 10 sampling methods, and 2 classifiers.
As described in Section \ref{classifiers}, the performance of the sampling strategies and ratios is measured based on the accuracy of the classifier that is trained on the resampled train set. 
We focus on the results obtained based on the RF classifier. The results based on the SVM classifier are in line with those of RF and are summarized in the corresponding tables and figures.

We begin by taking a close look at the effects of different sampling ratios based on the SMOTE sampling  algorithm. In Table \ref{smote}, we provide the accuracy results, as measured on the test set, for the RF classifier that is trained on a partially balanced set using the SMOTE algorithm.
The first column in the table, labeled \textsf{orig}, shows the accuracy on the original imbalanced data. The last column in the table, labeled \textsf{max}, shows the maximum accuracy over all sampling ratios. Thus, for the \textit{ecoli} dataset the maximum accuracy is 0.8473 which is achieved at the sampling ratio $r=0.8$. Similarly, for the \textit{abalone} dataset, the maximum accuracy is 0.6698 which is achieved at the sampling ratio $r=0.9$. Note that almost all the values in the \textsf{max} column are greater than the corresponding values in the column $r=1$ which indicates that maximum accuracy is rarely achieved with full resampling. Indeed, it can be seen from Table \ref{smote} that the maximum accuracy is often achieved at lower sampling ratios between $r=0.2$ and  $r= 0.9$. 
It can also be seen from  Table \ref{smote} that the accuracy generally increases as the sampling ratio increases from $r=0.2$ to $r=1$. This pattern holds across all the datasets in the table. Furthermore, partial resampling, at any class ratio, produces higher accuracy than the original imbalanced data.

\begin{table}[htb]
\centering
\caption{Balanced accuracy using the SMOTE sampling algorithm for the RF classifier.}
\label{smote}
\scalebox{0.9}{
\begin{tabular}{lrrrrrrrrrrr}
\toprule
{} &    orig &     0.2 &     0.3 &     0.4 &     0.5 &     0.6 &     0.7 &     0.8 &     0.9 &       1 &  max \\
\midrule
ecoli          &  0.7343 &  0.7677 &  0.7843 &  0.8070 &  0.8376 &  0.8123 &  0.8204 &  \bf{0.8473} &  0.8326 &  0.8057 &        0.8473 \\
abalone        &  0.5434 &  0.5768 &  0.6067 &  0.6267 &  0.6421 &  0.6569 &  0.6661 &  0.6619 &  \bf{0.6698} &  0.6585 &        0.6698 \\
car\_eval\_34    &  0.9366 &  0.9644 &  0.9609 &  0.9534 &  0.9766 &  0.9682 &  0.9614 &  0.9605 &  \bf{0.9791} &  0.9774 &        0.9791 \\
libras\_move    &  0.6964 &  0.7771 &  0.8128 &  0.8307 &  0.8557 &  0.8557 &  \bf{0.9027} &  0.8557 &  0.8807 &  0.8557 &        0.9027 \\
spectrometer   &  0.8312 &  0.8408 &  0.8806 &  0.8902 &  0.8902 &  \bf{0.9205} &  0.8989 &  0.9075 &  0.8882 &  0.8991 &        0.9205 \\
solar\_flare\_m0 &  0.5175 &  0.5390 &  0.5418 &  0.5492 &  0.5436 &  \bf{0.5578} &  0.5409 &  0.5560 &  0.5290 &  0.5487 &        0.5578 \\
car\_eval\_4     &  0.9118 &  0.9328 &  0.9226 &  0.9887 &  0.9890 &  0.9631 &  0.9606 &  0.9632 &  \bf{0.9916} &  0.9896 &        0.9916 \\
oil            &  0.6658 &  0.7236 &  0.7291 &  0.7158 &  0.7464 &  \bf{0.7574} &  0.7297 &  0.7381 &  0.7308 &  0.7012 &        0.7574 \\
sick\_euthyroid &  0.9184 &  0.9311 &  0.9335 &  0.9333 &  \bf{0.9380} &  0.9342 &  0.9362 &  0.9370 &  0.9315 &  0.9326 &        0.9380 \\
wine\_quality   &  0.5775 &  0.6511 &  0.6485 &  0.6540 &  0.6629 &  0.6595 &  0.6735 &  0.6726 &  \bf{0.6774} &  0.6557 &        0.6774 \\
pen\_digits     &  0.9805 &  0.9871 &  0.9860 &  0.9861 &  \bf{0.9889} &  0.9885 &  0.9875 &  0.9880 &  0.9888 &  0.9879 &        0.9889 \\
arrhythmia     &  0.5000 &  0.5000 &  0.5000 &  0.5196 &  0.5178 &  0.5155 &  0.5542 &  \bf{0.5554} &  0.5155 &  0.5542 &        0.5554 \\
satimage       &  0.7504 &  0.7831 &  0.8000 &  0.8065 &  0.8084 &  0.8054 &  0.8082 &  0.8148 &  \bf{0.8196} &  0.8146 &        0.8196 \\
us\_crime       &  0.6729 &  0.7165 &  0.7266 &  0.7430 &  0.7392 &  0.7279 &  0.7423 &  0.7474 &  \bf{0.7553} &  0.7537 &        0.7553 \\
thyroid\_sick   &  0.8778 &  0.8974 &  0.9175 &  0.9138 &  0.9206 &  0.9242 &  0.9249 &  \bf{0.9386} &  0.9312 &  0.9278 &        0.9386 \\
yeast\_ml8      &  0.4998 &  0.5000 &  \bf{0.5000} &  0.4993 &  0.4989 &  0.4986 &  0.4982 &  0.4967 &  0.4960 &  0.4978 &        0.5000 \\
optical\_digits &  0.8951 &  0.9074 &  0.9240 &  0.9240 &  0.9348 &  0.9309 &  0.9303 &  0.9357 &  0.9347 &  \bf{0.9386} &        0.9386 \\
scene          &  0.5179 &  0.5201 &  0.5350 &  0.5519 &  0.5533 &  0.5552 &  0.5510 &  0.5480 &  0.5496 &  \bf{0.5615} &        0.5615 \\
coil\_2000      &  0.5230 &  0.5305 &  0.5305 &  0.5317 &  0.5326 &  0.5327 &  0.5331 &  0.5332 &  \bf{0.5348} &  0.5322 &        0.5348 \\
isolet         &  0.7744 &  0.8611 &  0.8851 &  0.8933 &  0.8969 &  0.9021 &  0.9004 &  0.9050 &  \bf{0.9079} &  0.9045 &        0.9079 \\
\bottomrule
\end{tabular}}
\end{table}

\begin{figure}[H]
\center
\includegraphics[width=0.8\textwidth]{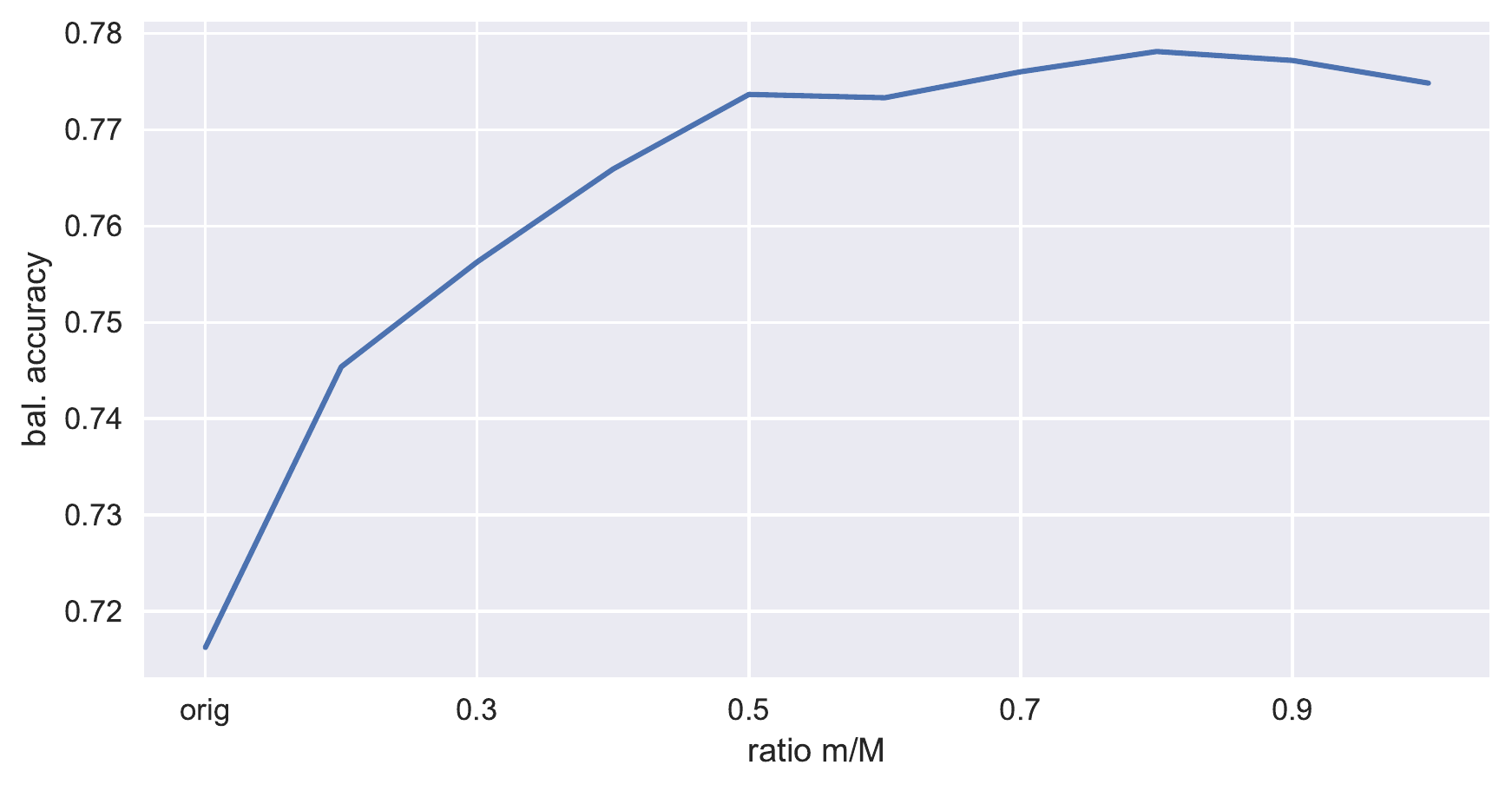}
\caption{The average accuracy at each class ratio using the SMOTE algorithm, where the average is taken over all the datasets based on Table \ref{smote}.}
\label{smote_fig}
\end{figure}

In Figure \ref{smote_fig}, we present the average accuracy at each sampling ratio calculated over all the datasets based on Table \ref{smote}. It can be seen from Figure \ref{smote_fig} that the greatest average accuracy occurs at $r=0.8$. In particular, the mean accuracy at $r=0.8$ is greater than the  accuracy at $r=1$. We also observe that the accuracy generally increases as the class ratio is increased via sampling.

\begin{figure}[H]
\center
\includegraphics[width=1\textwidth]{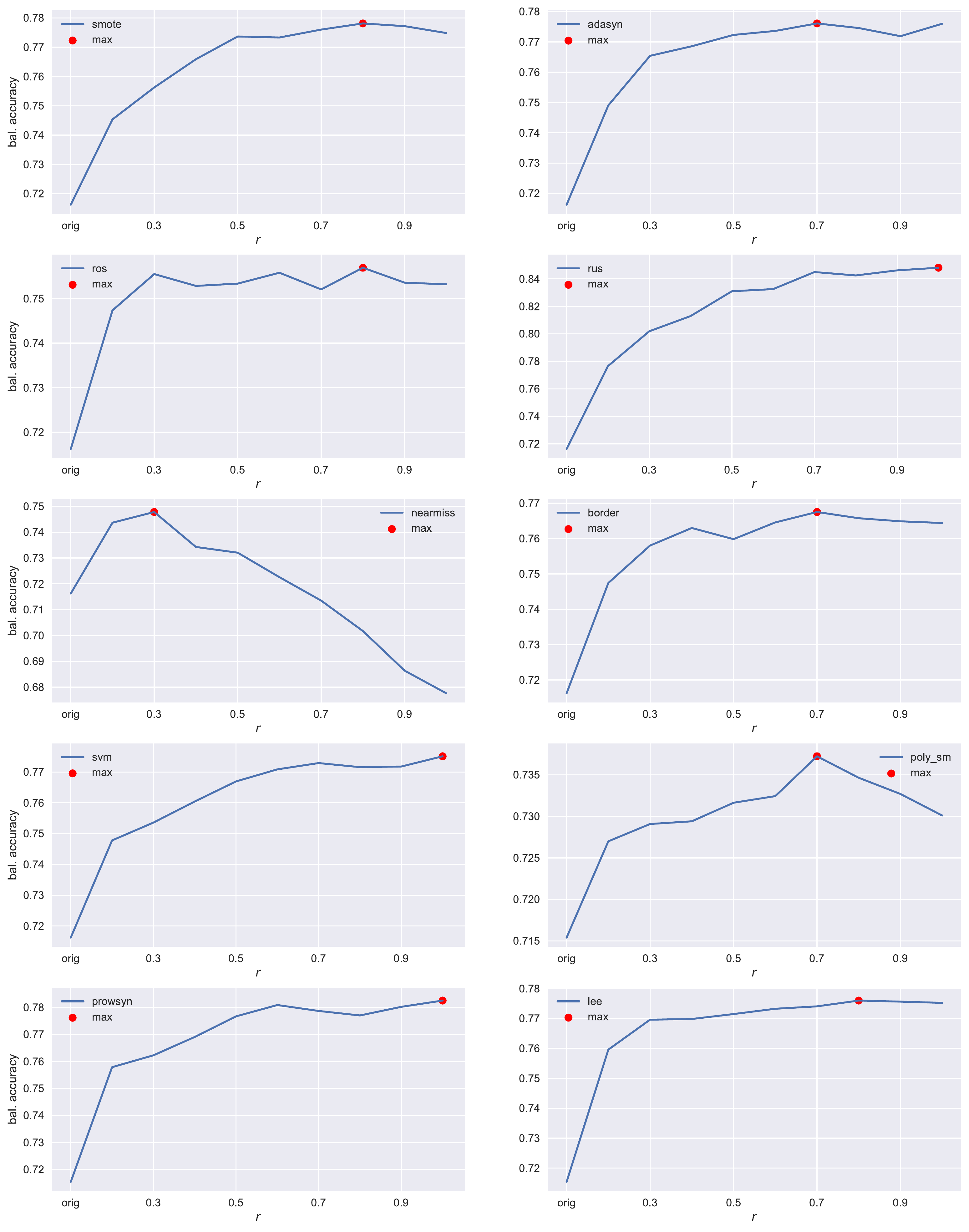}
\caption{The mean accuracy of each sampling method, where the average is taken over all the datasets in Table \ref{datasets}.
The accuracy is measured by the performance of the RF classifier trained on the sampled data.}
\label{comparison}
\end{figure}

We extend the above analysis that was done for the SMOTE algorithm to other sampling methods.
Figure \ref{comparison} shows the mean accuracy for each sampling method contained in Table \ref{sampling}. The mean is taken over all datasets in Table \ref{datasets}. As discussed earlier, the SMOTE algorithm achieves the best accuracy at ratio $r=0.8$ (Figure \ref{comparison}a). For the ADASYN method, the best result is obtained at $r=0.7$ and $r=1$ (Figure \ref{comparison}b). For the ROS, NearMiss, Border, PolySM, and Lee methods the highest accuracy is achieved at sampling ratios of $r=0.8, r=0.3, r=0.7, r=0.7$, and $r=0.8$ respectively. The results show that on average the best performance of these algorithms is achieved at $r<1$. 

We also note that in most cases, sampling has a positive effect on accuracy. As shown in Figure \ref{comparison}, the  accuracy at \textsf{orig} is lower than at all other ratios. This pattern holds for all the sampling methods. Based on Figure  \ref{comparison}, we conclude, that while data sampling has a significant positive effect on the accuracy of the classifier, the optimal sampling ratio is generally below the full resampling. In fact, our experiments show that the best sampling ratio of minority to majority classes is between $r=0.7$ and $r=0.8$.

In Table \ref{mean_sampling}, we provide yet another perspective on the performance of different sampling ratios. In particular, we present the average accuracy for each combination of dataset and sampling ratio, where for each pair (dataset, sampling ratio) the average is taken over all the sampling methods in Table \ref{sampling}. 
As shown in Table \ref{mean_sampling}, the highest mean accuracy - taken over all the sampling methods - for the \textit{ecoli} dataset is 0.8174 which is achieved at the sampling ratio $r=0.7$. Similarly, for the \textit{car\_eval\_34} dataset, the highest average accuracy is 0.9667 which is achieved at $r=0.5$. As can be seen from Table \ref{mean_sampling}, in most cases, the highest average accuracy occurs at $r<1$. Only in 4 out of 20 datasets the best accuracy is achieved at $r=1$ . We also observe that the original sampling ratio yields the lowest average accuracy. 
Based on Table \ref{mean_sampling}, we conclude that while resampling increases the average balanced accuracy for each dataset, the best results are often obtained at resampling ratios less than $r=1$. The practical implications of the accuracy results are that
\begin{enumerate}
\item[i.] data sampling always improves the accuracy of the classifier, 
\item[ii.] the best sampling ratio is often between $r=0.7$ and $r=0.8$, and 
\item[iii.] to obtain the best results it is necessary to perform a grid search evaluation of various sampling ratios.
\end{enumerate}

\begin{table}[htb]
\centering
\caption{The mean balanced accuracy on each dataset, where the average is calculated over all the sampling methods.}
\label{mean_sampling}
\scalebox{0.9}{
\begin{tabular}{lrrrrrrrrrrr}
\toprule
{} &    orig &     0.2 &     0.3 &     0.4 &     0.5 &     0.6 &     0.7 &     0.8 &     0.9 &       1 &  max \\
\midrule
ecoli          &  0.7343 &  0.7649 &  0.8019 &  0.7957 &  0.7957 &  0.8033 &  \bf{0.8174} &  0.8042 &  0.7848 &  0.7798 &        0.8174 \\
abalone        &  0.5434 &  0.5766 &  0.5911 &  0.5956 &  0.6010 &  0.6106 &  0.6158 &  0.6218 &  0.6271 &  \bf{0.6280} &        0.6280 \\
car\_eval\_34    &  0.9366 &  0.9560 &  0.9579 &  0.9592 &  \bf{0.9667} &  0.9615 &  0.9643 &  0.9561 &  0.9578 &  0.9541 &        0.9667 \\
libras\_move    &  0.6964 &  0.7992 &  0.8259 &  0.8302 &  0.8501 &  0.8541 &  \bf{0.8614} &  0.8471 &  0.8373 &  0.8494 &        0.8614 \\
spectrometer   &  0.8312 &  0.8663 &  0.8885 &  0.8805 &  0.8913 &  \bf{0.9016} &  0.8842 &  0.8876 &  0.8862 &  0.8821 &        0.9016 \\
solar\_flare\_m0 &  0.5175 &  0.5518 &  0.5526 &  0.5592 &  0.5612 &  0.5601 &  0.5607 &  0.5614 &  0.5575 &  \bf{0.5672} &        0.5672 \\
car\_eval\_4     &  0.9118 &  0.9373 &  0.9467 &  0.9570 &  0.9531 &  0.9548 &  0.9516 &  0.9499 &  \bf{0.9588} &  0.9468 &        0.9588 \\
oil            &  0.6658 &  0.7224 &  0.7306 &  0.7296 &  0.7331 &  \bf{0.7388} &  0.7343 &  0.7296 &  0.7201 &  0.7161 &        0.7388 \\
sick\_euthyroid &  0.9184 &  \bf{0.9263} &  0.9228 &  0.9189 &  0.9185 &  0.9174 &  0.9164 &  0.9156 &  0.9151 &  0.9122 &        0.9263 \\
wine\_quality   &  0.5775 &  0.6545 &  0.6651 &  0.6674 &  0.6690 &  0.6710 &  \bf{0.6711} &  0.6661 &  0.6675 &  0.6625 &        0.6711 \\
pen\_digits     &  0.9805 &  0.9855 &  0.9862 &  0.9863 &  \bf{0.9870} &  0.9869 &  0.9804 &  0.9814 &  0.9813 &  0.9817 &        0.9870 \\
arrhythmia     &  0.5000 &  0.5307 &  0.5287 &  0.5340 &  0.5492 &  0.5434 &  0.5650 &  0.5700 &  0.5705 &  \bf{0.5772} &        0.5772 \\
satimage       &  0.7504 &  0.7879 &  0.7895 &  0.7916 &  0.7934 &  0.7958 &  0.7963 &  0.7970 &  \bf{0.7972} &  0.7971 &        0.7972 \\
us\_crime       &  0.6729 &  0.7121 &  0.7287 &  0.7330 &  0.7375 &  0.7354 &  0.7381 &  0.7387 &  0.7362 &  \bf{0.7383} &        0.7387 \\
thyroid\_sick   &  0.8778 &  0.8989 &  0.9069 &  \bf{0.9083} &  0.9053 &  0.9069 &  0.9021 &  0.9021 &  0.8986 &  0.8964 &        0.9083 \\
yeast\_ml8      &  0.4998 &  0.5000 &  0.5000 &  0.5017 &  0.5060 &  0.5057 &  \bf{0.5118} &  0.5100 &  0.5082 &  0.5083 &        0.5118 \\
optical\_digits &  0.8951 &  0.9143 &  0.9208 &  0.9257 &  \bf{0.9296} &  0.9258 &  0.9262 &  0.9261 &  0.9239 &  0.9242 &        0.9296 \\
scene          &  0.5179 &  0.5312 &  0.5460 &  0.5559 &  0.5616 &  0.5610 &  0.5639 &  0.5645 &  0.5656 &  \bf{0.5667} &        0.5667 \\
coil\_2000      &  0.5230 &  0.5372 &  0.5394 &  0.5413 &  0.5432 &  0.5439 &  0.5451 &  \bf{0.5457} &  0.5454 &  0.5445 &        0.5457 \\
isolet         &  0.7692 &  0.8503 &  0.8696 &  0.8815 &  \bf{0.8854} &  0.8818 &  0.8801 &  0.8827 &  0.8819 &  0.8817 &        0.8854 \\
\bottomrule
\end{tabular}}
\end{table}

The results in Table \ref{mean_sampling} show that there is little relation between the optimal sampling ratio and dataset characteristics. In particular, the original imbalance ratio of a dataset plays little role in determining its optimal sampling ratio. For instance, the \textit{ecoli} and \textit{wine\_quality} datasets which have the lowest and highest imbalance ratios (Table \ref{datasets}) respectively, both achieve the optimal performance at the sampling ratio $r=0.7$ (Table \ref{mean_sampling}). Similarly, the number of features in the dataset has little effect on the optimal sampling ratio. For instance, the datasets \textit{car\_eval\_34} and \textit{isolet} both have the same optimal sampling ratio $r=0.5$, but the number of features is 21 and 617, respectively.

The only meaningful relationship between the optimal sampling ratio and the properties of the dataset that can be derived from the results in Table \ref{mean_sampling} is with respect to the number of samples. In particular, datasets with larger number of samples tend to have smaller optimal ratio. To derive this relationship, we consider the six datasets with the optimal ratio $r\leq0.5$ : \textit{optical\_digits}, \textit{pen\_digits},  \textit{sick\_euthroid}, \textit{car\_eval\_34}, \textit{isolet}, and \textit{thyroid\_sick}. The average number of samples in the datasets with $r\leq0.5$ is 5512, while the average number of samples in the datasets with $r\geq 0.6$ is 2705. Thus, the average number of samples in datasets with $r\leq0.5$  is twice larger than in  the datasets with $r\geq 0.6$. We postulate that given a large number of samples in the original dataset, there is less need for additional minority samples to learn the patterns within data.
Conversely, a dataset with relatively few samples requires more minority samples for classifier to properly learn its patterns.

The mean F1-macro scores are presented in Table \ref{f1-macro}. 
As above, the average is taken over all the sampling methods. The results indicate that the optimal F1-macro values are achieved at lower sampling ratios. In particular, the highest F1-macro often occurs at $r=0.2$ and $r=0.3$. It indicates that to obtain the best performance in terms of precision and recall we need to apply a low sampling ratio. We also note that full sampling  at $r=1$ produces suboptimal results which supports our findings based on the balanced accuracy metric.

\begin{table}[htb]
\centering
\caption{The mean F1-macro score on each dataset, where the average is calculated over all the sampling methods.}
\label{f1-macro}
\scalebox{0.9}{
\begin{tabular}{lrrrrrrrrrrr}
\toprule
{} &    orig &     0.2 &     0.3 &     0.4 &     0.5 &     0.6 &     0.7 &     0.8 &     0.9 &       1 &  max \\
\midrule
ecoli          &  0.7740 &  0.7850 & \bf{0.8009} &  0.7853 &  0.7768 &  0.7734 &  0.7776 &  0.7604 &  0.7433 &  0.7295 &        0.8009 \\
abalone        &  0.5551 &  0.5874 &  \bf{0.5952} &  0.5917 &  0.5875 &  0.5881 &  0.5831 &  0.5818 &  0.5809 &  0.5783 &        0.5952 \\
car\_eval\_34    &  0.9494 &  0.9529 &  0.9523 &  0.9510 &  \bf{0.9535} &  0.9487 &  0.9436 &  0.9340 &  0.9316 &  0.9240 &        0.9535 \\
libras\_move    &  0.7605 &  0.8330 &  0.8401 &  0.8446 &  0.8538 &  \bf{0.8568} &  0.8556 &  0.8398 &  0.8327 &  0.8384 &        0.8568 \\
spectrometer   &  0.8685 &  0.8804 &  \bf{0.8856} &  0.8714 &  0.8757 &  0.8773 &  0.8610 &  0.8590 &  0.8498 &  0.8508 &        0.8856 \\
solar\_flare\_m0 &  0.5209 &  \bf{0.5412} &  0.5330 &  0.5328 &  0.5232 &  0.5141 &  0.5136 &  0.5132 &  0.5014 &  0.5079 &        0.5412 \\
car\_eval\_4     &  0.9450 &  0.9413 &  0.9442 &  \bf{0.9473} &  0.9413 &  0.9249 &  0.9178 &  0.9109 &  0.9141 &  0.9078 &        0.9473 \\
oil            &  0.7239 &  0.7272 &  \bf{0.7279} &  0.7258 &  0.7162 &  0.7227 &  0.7144 &  0.7102 &  0.6977 &  0.6941 &        0.7279 \\
sick\_euthyroid &  \bf{0.9303} &  0.9221 &  0.9043 &  0.8960 &  0.8931 &  0.8893 &  0.8869 &  0.8853 &  0.8825 &  0.8805 &        0.9303 \\
wine\_quality   &  0.6136 &  \bf{0.6599} &  0.6576 &  0.6485 &  0.6414 &  0.6359 &  0.6308 &  0.6219 &  0.6188 &  0.6105 &        0.6599 \\
pen\_digits     &  0.9888 &  0.9913 &  \bf{0.9917} &  0.9911 &  0.9914 &  0.9886 &  0.9684 &  0.9700 &  0.9687 &  0.9697 &        0.9917 \\
arrhythmia     &  0.4858 &  0.5234 &  0.5203 &  0.5212 &  0.5297 &  0.5252 &  \bf{0.5424} &  0.5364 &  0.5345 &  0.5359 &        0.5424 \\
satimage       &  0.8016 &  \bf{0.8169} &  0.7925 &  0.7818 &  0.7777 &  0.7752 &  0.7697 &  0.7654 &  0.7624 &  0.7578 &        0.8169 \\
us\_crime       &  0.7240 &  \bf{0.7344} &  0.7308 &  0.7222 &  0.7189 &  0.7099 &  0.7094 &  0.7050 &  0.6979 &  0.6982 &        0.7344 \\
thyroid\_sick   &  \bf{0.9094} &  0.8966 &  0.8946 &  0.8902 &  0.8856 &  0.8845 &  0.8748 &  0.8715 &  0.8669 &  0.8625 &        0.9094 \\
yeast\_ml8      &  0.4808 &  0.4809 &  0.4826 &  0.4851 &  \bf{0.4882} &  0.4821 &  0.4786 &  0.4715 &  0.4619 &  0.4579 &        0.4882 \\
optical\_digits &  0.9357 &  0.9477 &  0.9506 &  \bf{0.9523} &  0.9509 &  0.9374 &  0.9334 &  0.9293 &  0.9260 &  0.9240 &        0.9523 \\
scene          &  0.5158 &  0.5350 &  0.5487 &  0.5545 &  \bf{0.5573} &  0.5479 &  0.5469 &  0.5428 &  0.5370 &  0.5380 &        0.5573 \\
coil\_2000      &  0.5283 &  \bf{0.5380} &  0.5342 &  0.5295 &  0.5252 &  0.5212 &  0.5166 &  0.5125 &  0.5088 &  0.5041 &        0.5380 \\
isolet         &  0.8362 &  0.8923 &  \bf{0.8962} &  0.8946 &  0.8846 &  0.8738 &  0.8665 &  0.8658 &  0.8619 &  0.8586 &        0.8962 \\
\bottomrule
\end{tabular}}
\end{table}

The results in Tables \ref{mean_sampling} and \ref{f1-macro} show that the choice of the performance metric affects the optimal sampling ratio for imbalanced data. If the goal is to achieve the maximum accuracy then sampling ratios $r=0.7$ and $r=0.8$ often produce the best performance. On the other hand, sampling ratios $r=0.2$ and $r=0.3$ produce the best F1-macro values. These values can be used as default parameters in sampling methods. In addition, regardless of the performance metric, we observe that sampling usually improves the performance and full sampling rarely produces the best results. In general, we strongly recommend to perform a grid search over a range of sampling ratios using cross-validation to identify the optimal sampling ratio.

\begin{figure}[H]
\center
\includegraphics[width=1\textwidth]{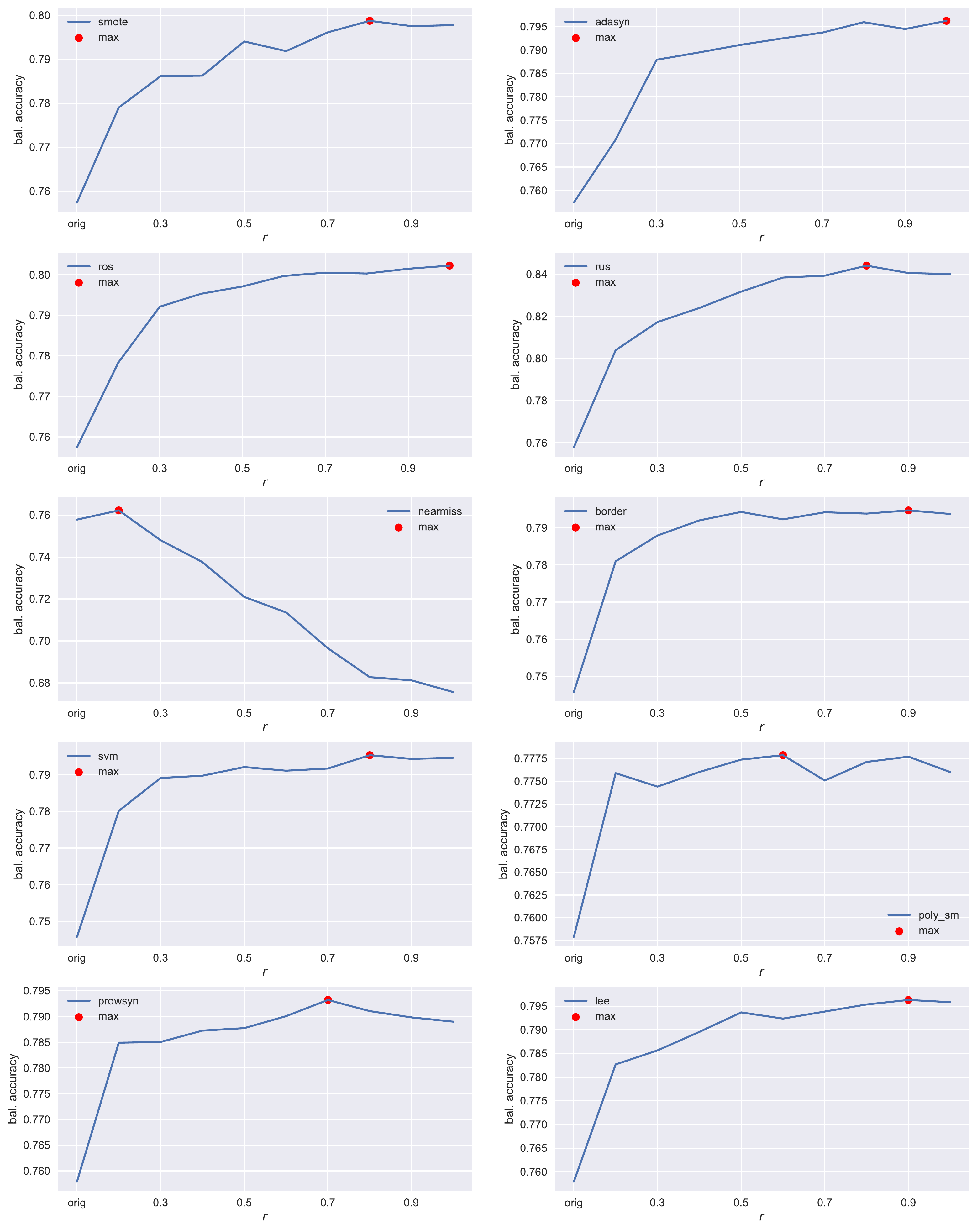}
\caption{The mean accuracy of the sampling methods as measured by the performance of the SVC classifier trained on the sampled data. The mean is taken over all the datasets in Table \ref{datasets}.}
\label{comparison_svc}
\end{figure}

The results using the SVM classifier are generally in line with above results obtained with the RF classifier. The SVM-based experiments show that sampling improves the accuracy of the classifier. The optimal sampling ratio often occurs at $r<1$. The main difference between the SVM and RF-based results is in the higher accuracy of the former. It appears that the SVM classifier is better suited for imbalanced data when used in conjunction with data sampling. The details of the SVM-based experiments are supplied in Table \ref{mean_sampling_svc} and Figure \ref{comparison_svc}.

As shown in  Table \ref{mean_sampling_svc}, there is no discernible relation between the optimal sampling ratio and the original imbalance ratio. Similarly, the number of features in the dataset does not seem to have much impact on the value of the optimal $r$. The only dataset property that appears to affect the optimal ratio is the number of samples. Datasets with more samples, on average, achieve the optimal performance at lower values of $r$. We conjecture that large datasets require fewer additional minority samples to have enough samples points for effective pattern recognition.

\begin{table}[htb]
\centering
\caption{The mean accuracy  on each dataset - as measured by the performance of the SVC classifier - calculated over all the sampling methods. }
\label{mean_sampling_svc}
\scalebox{0.9}{
\begin{tabular}{lrrrrrrrrrrr}
\toprule
{} &    orig &     0.2 &     0.3 &     0.4 &     0.5 &     0.6 &     0.7 &     0.8 &     0.9 &       1 &  max \\
\midrule
ecoli          &  0.8045 &  0.8205 &  0.8415 &  0.8331 &  \bf{0.8460} &  0.8281 &  0.8243 &  0.8173 &  0.8160 &  0.8107 &        0.8460 \\
abalone        &  0.5043 &  0.6312 &  0.6727 &  0.6904 &  0.6842 &  0.6851 &  0.6901 &  0.6942 &  0.6959 &  \bf{0.6978} &        0.6978 \\
car\_eval\_34    &  0.9641 &  0.9833 &  0.9844 &  0.9842 &  \bf{0.9858} &  0.9853 &  0.9834 &  0.9821 &  0.9837 &  0.9830 &        0.9858 \\
libras\_move    &  0.8768 &  0.8803 &  0.8849 &  0.8841 &  0.8880 &  0.8924 &  \bf{0.892}8 &  0.8909 &  0.8870 &  0.8885 &        0.8928 \\
spectrometer   &  0.8546 &  0.9076 &  0.9109 &  \bf{0.9192} &  0.9124 &  0.9163 &  0.9121 &  0.9040 &  0.9098 &  0.9058 &        0.9192 \\
solar\_flare\_m0 &  0.5058 &  0.5335 &  0.5417 &  0.5500 &  0.5566 &  0.5571 &  0.5607 &  0.5750 &  0.5718 &  \bf{0.5753} &        0.5753 \\
car\_eval\_4     &  0.9523 &  0.9874 &  0.9858 &  \bf{0.9968} &  0.9965 &  0.9927 &  0.9892 &  0.9847 &  0.9854 &  0.9870 &        0.9968 \\
oil            &  0.7468 &  0.7682 &  0.7726 &  0.7714 &  0.7744 &  \bf{0.7810} &  0.7750 &  0.7761 &  0.7706 &  0.7702 &        0.7810 \\
sick\_euthyroid &  0.8722 &  \bf{0.9101} &  0.9070 &  0.9061 &  0.9009 &  0.8955 &  0.8882 &  0.8915 &  0.8916 &  0.8794 &        0.9101 \\
wine\_quality   &  0.5694 &  0.6385 &  0.6489 &  0.6497 &  0.6551 &  0.6633 &  0.6670 &  0.6704 &  \bf{0.6720} &  0.6709 &        0.6720 \\
pen\_digits     &  0.9957 &  \bf{0.9957} &  0.9944 &  0.9944 &  0.9946 &  0.9956 &  0.9953 &  0.9955 &  0.9956 &  0.9955 &        0.9957 \\
arrhythmia     &  \bf{0.6385} &  0.5712 &  0.5579 &  0.5548 &  0.5535 &  0.5603 &  0.5596 &  0.5569 &  0.5491 &  0.5508 &        0.6385 \\
satimage       &  0.7725 &  0.8290 &  0.8383 &  0.8363 &  0.8442 &  0.8409 &  0.8429 &  0.8440 &  0.8437 &  \bf{0.8448} &        0.8448 \\
us\_crime       &  0.7044 &  0.7129 & \bf{0.7172} &  0.7147 &  0.7133 &  0.7132 &  0.7135 &  0.7120 &  0.7111 &  0.7099 &        0.7172 \\
thyroid\_sick   &  0.8115 &  0.8615 &  0.8712 &  \bf{0.8748} &  0.8709 &  0.8653 &  0.8693 &  0.8659 &  0.8668 &  0.8645 &        0.8748 \\
yeast\_ml8      &  0.4984 &  0.5005 &  0.5032 &  0.5062 &  0.5036 &  \bf{0.5106} &  0.5091 &  0.5068 &  0.5071 &  0.5082 &        0.5106 \\
optical\_digits &  0.9661 &  0.9703 &  0.9720 &  0.9722 &  0.9732 &  \bf{0.9732} &  0.9721 &  0.9720 &  0.9713 &  0.9690 &        0.9732 \\
scene          &  0.5399 &  0.5542 &  0.5556 &  0.5578 &  0.5619 &  0.5661 &  0.5648 &  \bf{0.5685} &  0.5667 &  0.5678 &        0.5685 \\
coil\_2000      &  0.5095 &  0.5283 &  0.5416 &  0.5475 &  0.5570 &  0.5608 &  0.5647 &  0.5652 &  0.5677 &  \bf{0.5680} &        0.5680 \\
isolet         &  0.9366 &  0.9466 &  0.9461 &  \bf{0.9501} &  0.9476 &  0.9396 &  0.9344 &  0.9350 &  0.9331 &  0.9297 &        0.9501 \\
\bottomrule
\end{tabular}}
\end{table}

As mentioned in the introduction, increasing the number of minority points through sampling has a dual effect on classifier performance. On one hand, sampling provides the classifier with more data to learn the representation of the minority class. On the other hand, since the new points are not generated from the true distribution, they may lead to model misspecification. As the the number of sampled points increases, the issues related to misspecification of the model begin to outweight the benefits of learning the minority class parameters. Our numerical experiments show that the point of inflection often occurs between $r=0.7$ and $r=0.8$. At this optimal ratio, we obtain enough information to produce statistically robust parameter estimates and avoid overly skewing the minority class distribution.

\begin{figure}[H]
\center
\includegraphics[width=1\textwidth]{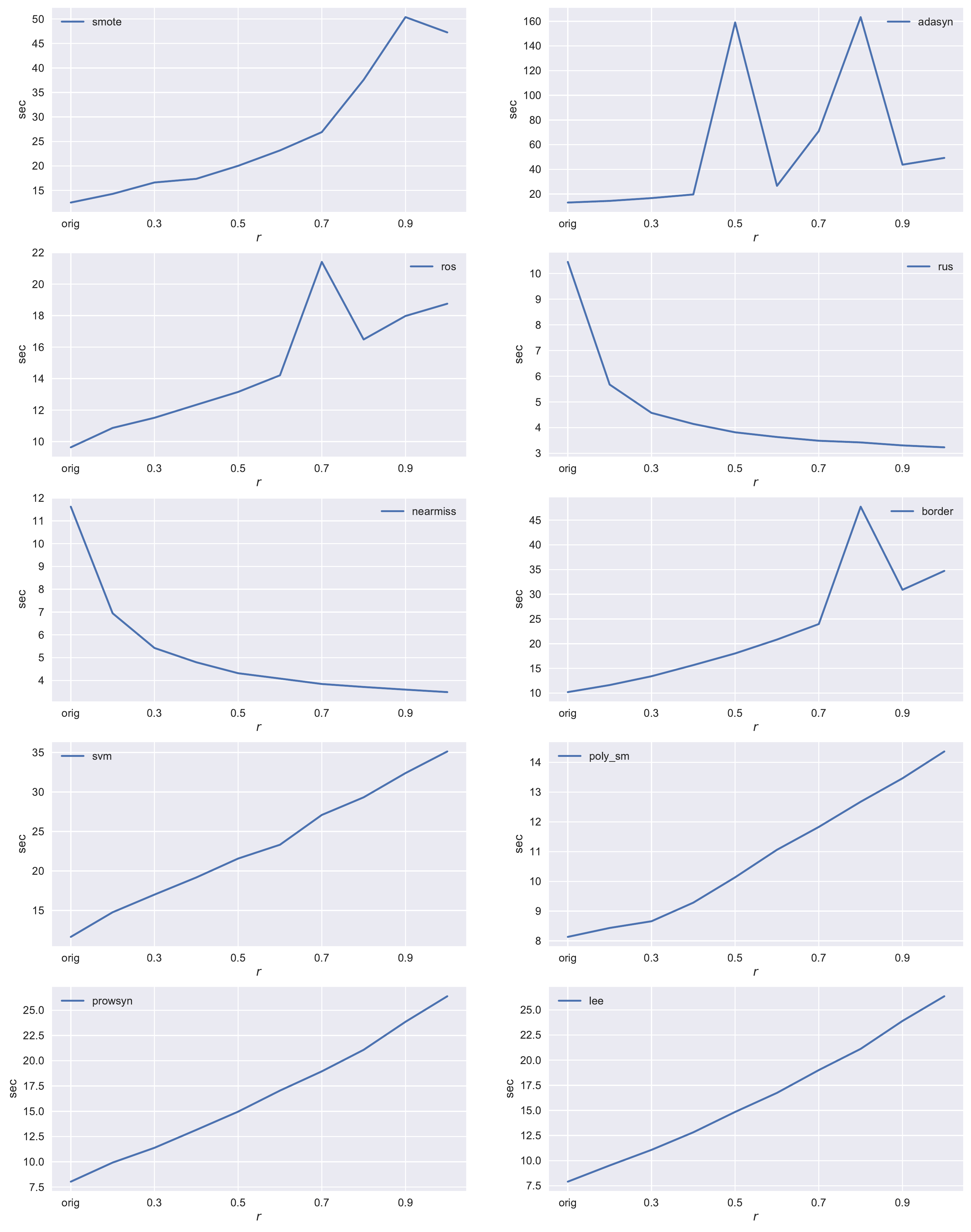}
\caption{The mean training times of the RF classifier on resampled data. The average is calculated over all the datasets in Table \ref{datasets}. The training time increases as the sampling ratio increases.}
\label{duration}
\end{figure}

Another important factor in the analysis of different sampling ratios is the classifier training times. The number of minority points in the dataset increases as the sampling ratio increases. For instance, given an imbalanced dataset with 1/10  original class ratio, the number of minority points increases 10-fold if we choose $r=1$ sampling ratio. The larger dataset leads to longer training times. The classifier training times are approximately linearly related to the size of the dataset. The training times for the RF classifier with different sampling ratios is supplied in Figure \ref{duration}.

In summary, our study reveals that while artificially increasing the class ratio in imbalanced data improves classification accuracy, full sampling ($r=1$) rarely produces the optimal results. Although the sampling ratio in the range $r=0.7$ to $r=0.8$ often produces the best accuracy and can be set as the default parameter value, it is best to conduct a grid search over a range of class ratios to identify the optimal ratio. Our study supports the previous findings in the literature about the general benefits of data sampling. However, the optimal sampling ratio identified in our experiments differs from the other studies.

The optimal sampling ratio depends on a number of factors including the data, sampling method, and performance metric. Given the same dataset, different sampling methods or performance measures can lead to different optimal ratios.
We find little relation between the optimal  ratio and dataset characteristics. In particular, dataset properties such as the original imbalance ratio and the number of features play a trivial role in determining the optimal ratio. The only factor that is found to affect the optimal ratio is the number of samples - datasets with large number of samples tend to have lower optimal ratio.
We conclude that the optimal sampling ratio depends uniquely on the distribution of the data points in the feature space for a given dataset.

%-----------------------------------------------------------------------------------------------------------------------------------------------------
%-----------------------------------------------------------------------------------------------------------------------------------------------------
\section{Conclusion}
Despite the existence of a large number sampling techniques for imbalanced data very few studies are dedicated to the issue of the optimal sampling ratio.
In this paper we fill the gap in the literature by conducting a large-scale study of the impact of sampling ratio on the classification accuracy. Our empirical study is based on 20 datasets and 10 sampling methods. The results show that i) sampling is generally a beneficial preprocessing step, ii) the optimal sampling ratio is between 0.7 and 0.8 albeit the exact value depends on the dataset, iii) full resampling ($r=1$) is rarely the best option, and iv) there is an inverse relation between the number of samples and the optimal ratio.
The present study enhances our understanding of the effects of the sampling ratio and provides insights into selecting the optimal ratio.

Our experiments show that while factors such the original imbalance ratio and the number of features do not play a significant role in determining the optimal ratio, the number of samples in the dataset may have a tangible impact. It is possible that there exists a more complex interplay between the data characteristics and the optimal ratio. Therefore, a future in-depth investigation into the relationship between the intrinsic data properties and the optimal sampling ratio is warranted.

\end{document}